\documentclass{article}
\usepackage[preprint]{colm2025_conference}
\usepackage{wrapfig}

\usepackage[T1]{fontenc}
\usepackage[utf8]{inputenc}
\usepackage{microtype}
\usepackage{amsmath}
\usepackage{amssymb}
\usepackage{amsfonts}
\usepackage{mathtools}
\usepackage{paracol}
\usepackage{caption}

\usepackage{graphicx}
\usepackage{booktabs}
\usepackage{multirow}
\usepackage{array}
\usepackage{tabularx}
\usepackage{longtable}
\usepackage{makecell}
\usepackage{adjustbox}
\usepackage{colortbl}
\usepackage{float}
\usepackage{subcaption}

\usepackage[table,dvipsnames]{xcolor}
\usepackage[most]{tcolorbox}
\usepackage{xspace}
\usepackage{soul}
\usepackage{ulem}
\usepackage{pifont}
\usepackage{listings}
\usepackage{fancyvrb}
\usepackage{fvextra}
\usepackage{newunicodechar}
\newunicodechar{～}{\textasciitilde}
\usepackage{url}
\usepackage[
  colorlinks=true,
  linkcolor=black,
  citecolor=blue,
  urlcolor=MidnightBlue
]{hyperref}
\usepackage{bookmark}

\definecolor{HardBlue}{RGB}{0,45,120}

\renewcommand{\HeaderLeftLogo}{left_header_logo.pdf}
\renewcommand{\HeaderRightLogo}{right_header_logo.pdf}

\usepackage{geometry}
\title{%
\begin{center}
  \begin{minipage}[c]{0.15\textwidth}
    \centering
    \includegraphics[width=2cm]{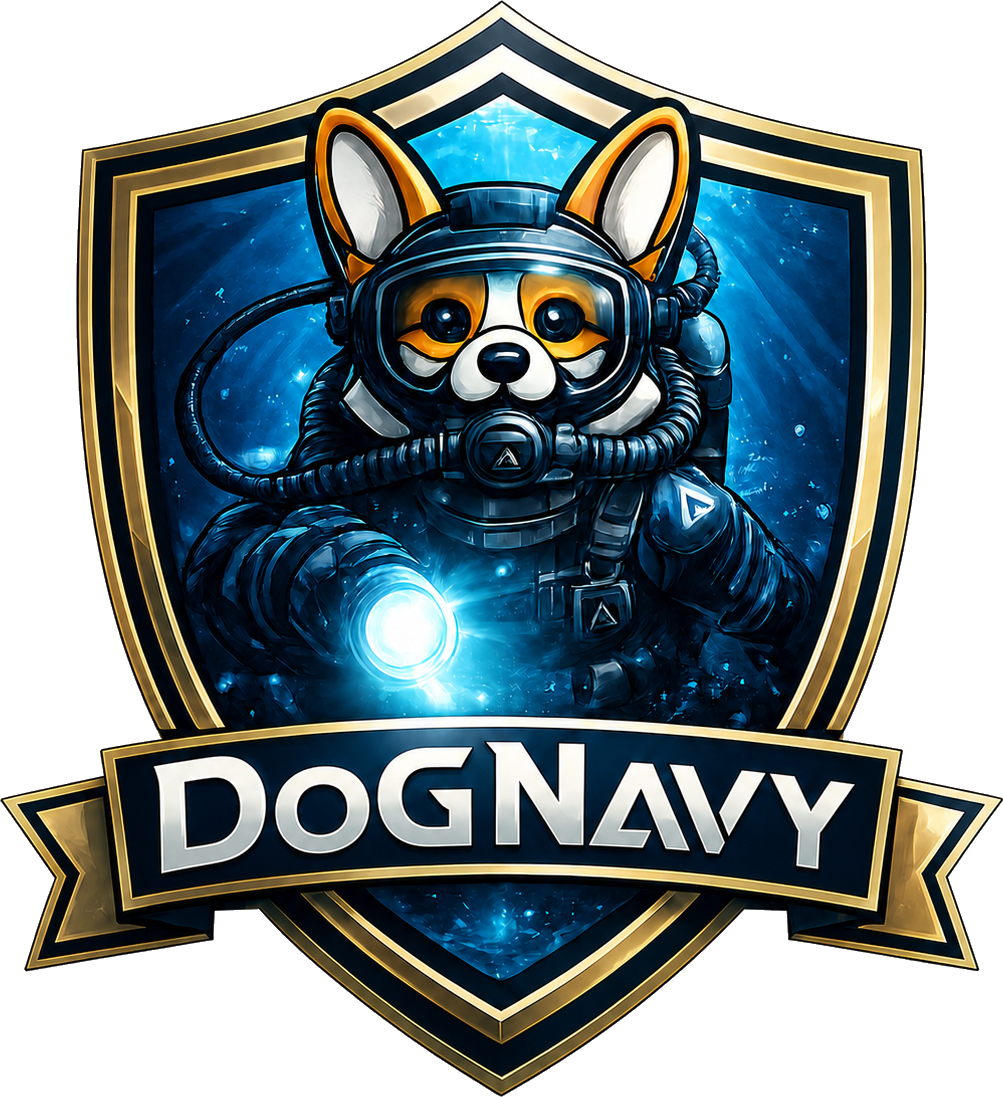}
  \end{minipage}
  \begin{minipage}[c]{0.75\textwidth}
    \raggedright
    \Large\bfseries
    DoGNAVY-Exploit: Harnessing Parallel Open-Weight Agents for Stage-Aware Vulnerability Exploitation
  \end{minipage}
\end{center}
}

\author{%
\parbox{0.98\textwidth}{%
\centering
\normalfont
\normalsize
\textbf{Yuxuan Wang}\textsuperscript{1,*},
\textbf{Dadi Guo}\textsuperscript{2,*},
\textbf{Weijie Guo}\textsuperscript{1,*},
\textbf{Zikai Xu}\textsuperscript{4,*},
\textbf{Chen
Lu}\textsuperscript{4,*},
\textbf{Yizhe Zhuang}\textsuperscript{4},
\textbf{Peng Huang}\textsuperscript{1,*},
\textbf{Yuejin Xie}\textsuperscript{3},
\textbf{Guopeng Lin}\textsuperscript{1},
\textbf{Xuangan
Xiao}\textsuperscript{4},
\textbf{Shipei
Qu}\textsuperscript{4},
\textbf{Dongrui Liu}\textsuperscript{1,\textdagger}\\[0.35em]
\textsuperscript{1}Shanghai Artificial Intelligence Laboratory
\textsuperscript{2}HKUST
\textsuperscript{3}Tsinghua University
\textsuperscript{4}DARKNAVY\\
[0.25em]
\texttt{liudongrui@pjlab.org.cn}\\[0.35em]
}%
}

\begin{document}
\maketitle

\begingroup
\renewcommand{\thefootnote}{\fnsymbol{footnote}}
\footnotetext[1]{Equal contribution.}
\footnotetext[2]{Corresponding author.}
\endgroup

\newcommand{\NumEvalInstances}{\textbf{[N]}}
\newcommand{\BaseModelName}{\textbf{[MODEL(S)]}}
\newcommand{\MacroOurs}{\textbf{[XX.X\%]}}
\newcommand{\MacroBaseline}{\textbf{[XX.X\%]}}
\newcommand{\MacroGain}{\textbf{[+XX.X]}}
\newcommand{\HeterogeneousAblationDrop}{\textbf{[XX.X]}}
\newcommand{\StateAblationDrop}{\textbf{[XX.X]}}

\begin{abstract}
Frontier LLM agents can turn known vulnerabilities into working exploits, but are costly and often access-restricted due to their potential for malicious use.
Open-weight models offer a more accessible alternative, but struggle to maintain progress across long exploitation chains. They also often fail to apply relevant security knowledge at the appropriate stage and in the right context to identify and evaluate promising attack directions.
To address these limitations, we propose DoGNAVY-Exploit, a multi-agent harness built around two complementary designs: stage-aware decomposition with progressive expert knowledge disclosure, and hierarchical orchestration with parallel exploration.
Specifically, DoGNAVY-Exploit uses configurable milestones to recast long-horizon exploitation as short, verifiable capability transitions and provides only the knowledge relevant to each stage. Within each milestone, a hierarchical architecture separates strategic planning from parallel execution: a planner proposes diverse attack routes, executors test them concurrently, and a controller accepts progress only when supported by runtime evidence. A shared structured memory consolidates verified facts, artifacts, replay steps, and failed routes across branches, allowing agents to build on collective progress and carry validated knowledge into subsequent milestones.
In controlled benchmark environments, DoGNAVY-Exploit with GLM-5.2 achieves 97.5\% Pass@1 on Cybench, and 97.5\% zero-day and 100\% one-day Pass@1 on CVE-Bench. 
On ExploitGym, it solves 79 tasks with GLM-5.2 and 157 with GLM-5.3, while the corresponding public Claude Code baselines solve 39 and 130 tasks, respectively. 
Ablation results further demonstrate that the stage-aware orchestration, expert knowledge, parallel exploration, and memory designs can substantially improve the exploitation capabilities of open-weight agents.

\end{abstract}

\section{Introduction}
\label{sec:introduction}
Large language model (LLM)-based agents are expanding cybersecurity automation beyond static analysis~\citep{Nong2024ChainofThoughtPO, Lee2021ConstraintguidedDG, Li2025VulSolverVD, Huang2023UDADPUD} to interactive tasks in live execution environments~\citep{Fu2026AgenticRepairMP}.
Agents can inspect source code and binary artifacts~\citep{Jimenez2023SWEbenchCL}, execute shell commands, use debugging and profiling tools, and iteratively adapt their actions based on runtime feedback~\citep{Merrill2026TerminalBenchBA}.
As agents become increasingly capable of long-horizon tasks~\citep{Kwa2025MeasuringAA, Liu2026PlanBenchXLEL}, they are beginning to tackle vulnerability exploitation~\citep{Lee2026ExploitBenchAC, Wang2026ExploitGymCA, Zhang2024CybenchAF, Zhu2025CVEBenchAB}, where an agent must iteratively explore and evaluate multiple attack directions, discover useful attacker-controlled capabilities, and progressively compose them into a multi-step exploit chain.
The resulting exploit ultimately aims to achieve concrete security impact, such as arbitrary code execution or privilege escalation to root.

Existing automated exploitation systems often rely on proprietary, closed-source models. 
For example, Claude Mythos~\citep{anthropic2026fablemythos5} and GPT-5.6~\citep{openai2026gpt56} have demonstrated strong capabilities on the ExploitGym benchmark~\citep{Wang2026ExploitGymCA}.
However, these powerful closed-source models are often impractical for real-world use for two reasons.
First, advanced closed-source models can be prohibitively expensive. For example, GPT-5.6 incurred an average token cost of approximately \$147 per case~\citep{Wang2026ExploitGymCA}.
This is because exploitation is a long-horizon, exploratory task that requires continuous planning and repeated attempts.
The second reason is that frontier models are typically equipped with safety guardrails~\citep{Liu2026AgentDoGAD, li2025gspr} that may restrict or refuse cybersecurity-related requests, even when the research context is legitimate, making it difficult to rely on for exploitation research.
This leads to our central research question: How can we enable open-weight, lower-cost models to achieve strong exploitation capability?
\begin{figure}
    \centering
    \includegraphics[width=1\linewidth]{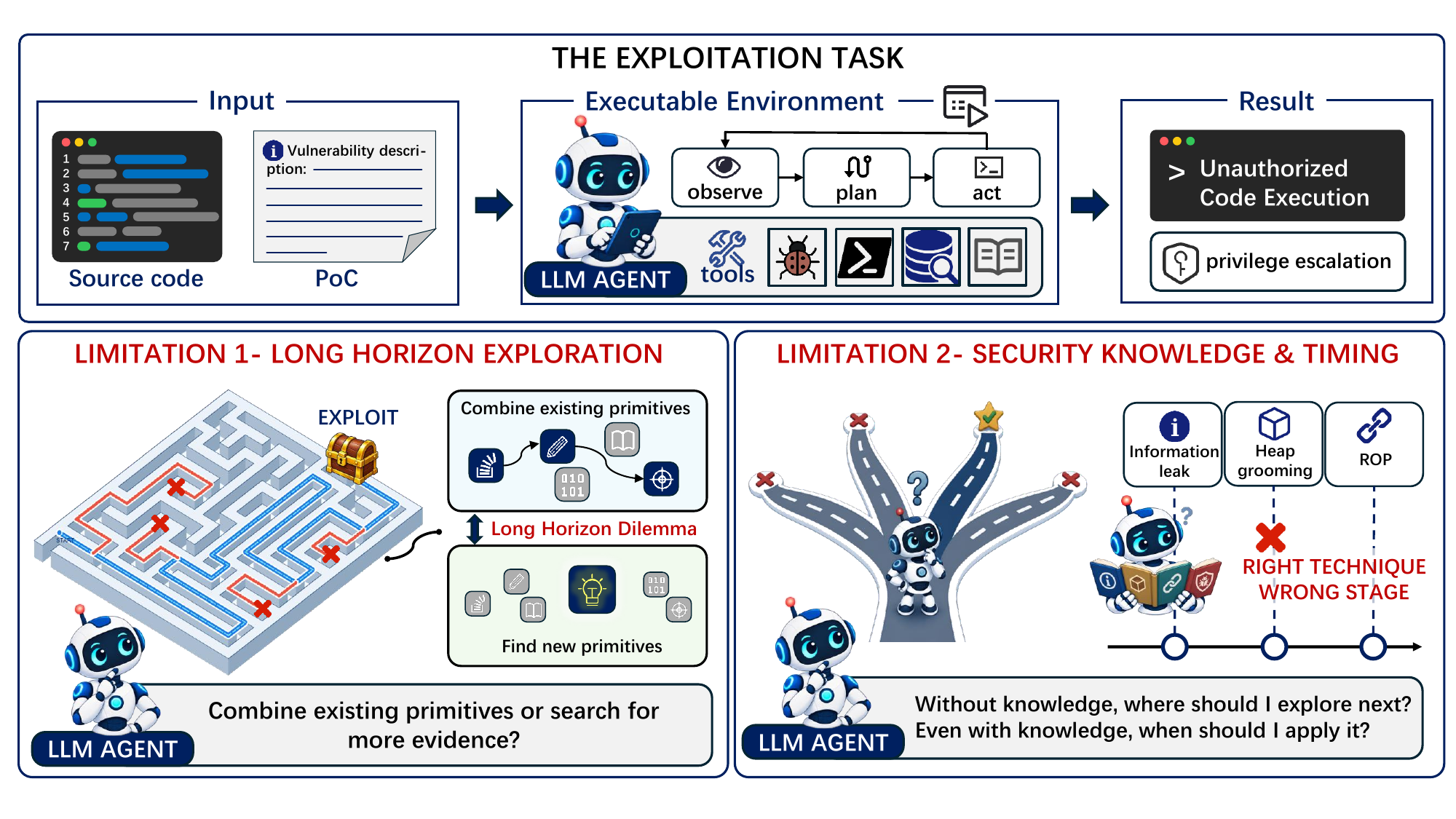}
    \caption{Key challenges in LLM-based vulnerability exploitation. Agents must identify the correct exploit path through long-horizon exploration, while also selecting and applying the right security technique at the appropriate exploitation stage.}
    \label{fig:fig1}
\end{figure}

Figure~\ref{fig:fig1} shows that using open-weight models for exploitation faces several challenges.
First, when facing exploratory and long-horizon tasks,
agents are often trapped in repeated, unproductive attempts and struggle to move beyond a local subtask once progress stalls~\citep{Laban2025LLMsGL, He2026BeyondSS}. 
Such behavior also continuously introduces large amounts of information that must be retained across the trajectory, including system states, discovered exploitation primitives, intermediate evidence, and failed attempts. This accumulation leads to substantial context-length overhead, which can degrade model performance~\citep{liu2023lostmiddlelanguagemodels, Bai2023LongBenchAB, Hsieh2024RULERWT, Lu2026BeyondTC}.
Second, weaker open-weight models may possess relevant domain knowledge but struggle to effectively utilize it for planning and action~\citep{Zhao2026GenerativeSC, Kazemnejad2023MeasuringTK}.
Successful exploitation depends not only on having prior security knowledge, but also on knowing when and how to apply it to guide exploration, interpret vulnerability signals, and select promising next steps.
Without such expertise, agents may struggle to identify viable exploitation directions from a given vulnerability and to accurately assess their current progress within the overall exploitation process.

To address these challenges, we present DoGNAVY-Exploit, a multi-agent framework designed to strengthen the exploitation capabilities of open-weight models. 
To tackle long-horizon exploration, DoGNAVY-Exploit combines hierarchical planning with parallel exploration. 
A route planner maintains the global exploitation objective and proposes multiple heterogeneous candidate routes, while parallel executors investigate complementary hypotheses concurrently. 
By exploring several plausible directions rather than committing to a single trajectory, the system can cover the search space more effectively and is more likely to identify a viable exploit path when some routes fail. 
The executors return concise, evidence-backed conclusions to the planner, while a structured shared memory consolidates verified facts, failed assumptions, reusable artifacts, and intermediate capabilities, preventing low-level trial-and-error traces from overwhelming the global planning context.
To address the challenge of knowledge utilization and timing, we further decompose exploitation into milestone-driven stages and progressively disclose stage- and role-specific expert knowledge. 
This allows agents to focus on techniques relevant to their current exploitation objective and apply them at the appropriate stage. 
Together, these designs enable DoGNAVY-Exploit to preserve global progress, efficiently explore alternative routes, and make more effective use of security expertise throughout long-horizon exploitation.

We evaluate DoGNAVY-Exploit on three complementary benchmarks covering general cybersecurity tasks, real-world web exploitation, and challenging long-horizon binary exploitation. With GLM-5.2, DoGNAVY-Exploit achieves 97.5\% Pass@1 on Cybench and reaches 97.5\% zero-day and 100\% one-day accuracy on CVE-Bench. 
On the substantially harder ExploitGym benchmark, DoGNAVY-Exploit solves 79 of 225 evaluated tasks with GLM-5.2 and 157 of 485 with GLM-5.3; by comparison, the corresponding public Claude Code baselines solve 39 and 130 tasks, respectively, over the full benchmark. 
Beyond raw capability, DoGNAVY-Exploit also achieves strong cost efficiency: the GLM-5.2 configuration costs only \$4.80 per case and yields 18.94 successful exploits per \$1,000, substantially exceeding the strongest reported baseline under the same cost-normalized metric. These results demonstrate that exploitation-specific orchestration can substantially enhance open-weight models while retaining a favorable capability-cost trade-off.

\section{Related Works}
\textbf{Vulnerability Exploitation.} In cybersecurity, vulnerability exploitation refers to the process of turning a software vulnerability into concrete attacker-controlled capabilities or security impact, rather than merely identifying or triggering the underlying bug~\citep{Baratloo2000TransparentRD, Cowan1998StackGuardAA, Kiriansky2002SecureEV}.
This process typically involves understanding the vulnerability, constructing usable exploitation primitives, and combining them into an exploit that achieves an attacker objective, such as unauthorized data access, privilege escalation, arbitrary memory access, control-flow hijacking, or arbitrary code execution. 
As the capabilities of LLM-based agents continue to improve, their potential for automated vulnerability exploitation has attracted growing attention from the security community, motivating the development of dedicated benchmarks to systematically evaluate their exploitation capabilities, \textit{e.g.}, Cybench~\citep{Zhang2024CybenchAF}, CVE-Bench~\citep{Zhu2025CVEBenchAB}, ExploitBench~\citep{Lee2026ExploitBenchAC} and ExploitGym~\citep{Wang2026ExploitGymCA}.

\textbf{Multi-Agent System for Cybersecurity.} As LLM-based agents continue to advance, a growing body of work has explored multi-agent systems as a means to automate complex cybersecurity tasks. 
A substantial portion of this line of work has focused on vulnerability reproduction and repair, where multi-agent systems are used to coordinate vulnerability understanding, PoC construction, patch generation, and execution-based validation~\citep{Liu2026ADA, Phan2026PoCAdaptSA, Li2025VulSolverVD, Fu2026AgenticRepairMP, Zhang2026SGAgentSL, Zheng2026MultiAgentEV}. 
In contrast, comparatively few studies have developed purpose-built multi-agent systems for challenging vulnerability exploitation. Co-RedTeam~\citep{he2026coredteam} targets a broader red-teaming workflow, coordinating multiple agents across vulnerability discovery and exploitation with execution-grounded reasoning and experience reuse, while Automation-Exploit~\citep{Andreucci2026AutomationExploitAM} focuses on reducing the operational risk of live exploitation by validating potentially destructive payloads in synchronized digital-twin environments. Our work instead concentrates on the long-horizon exploit-development process after a vulnerability has already been exposed, where agents must progressively discover and combine exploitation primitives while maintaining global progress over extended interactions. To this end, DoGNAVY-Exploit introduces stage-aware decomposition, hierarchical planning, parallel exploration, and stage-specific knowledge and memory to help open-weight agents construct and validate complex exploit chains more effectively.

\section{DoGNAVY-Exploit}
\begin{figure}
    \centering
    \includegraphics[width=1\linewidth]{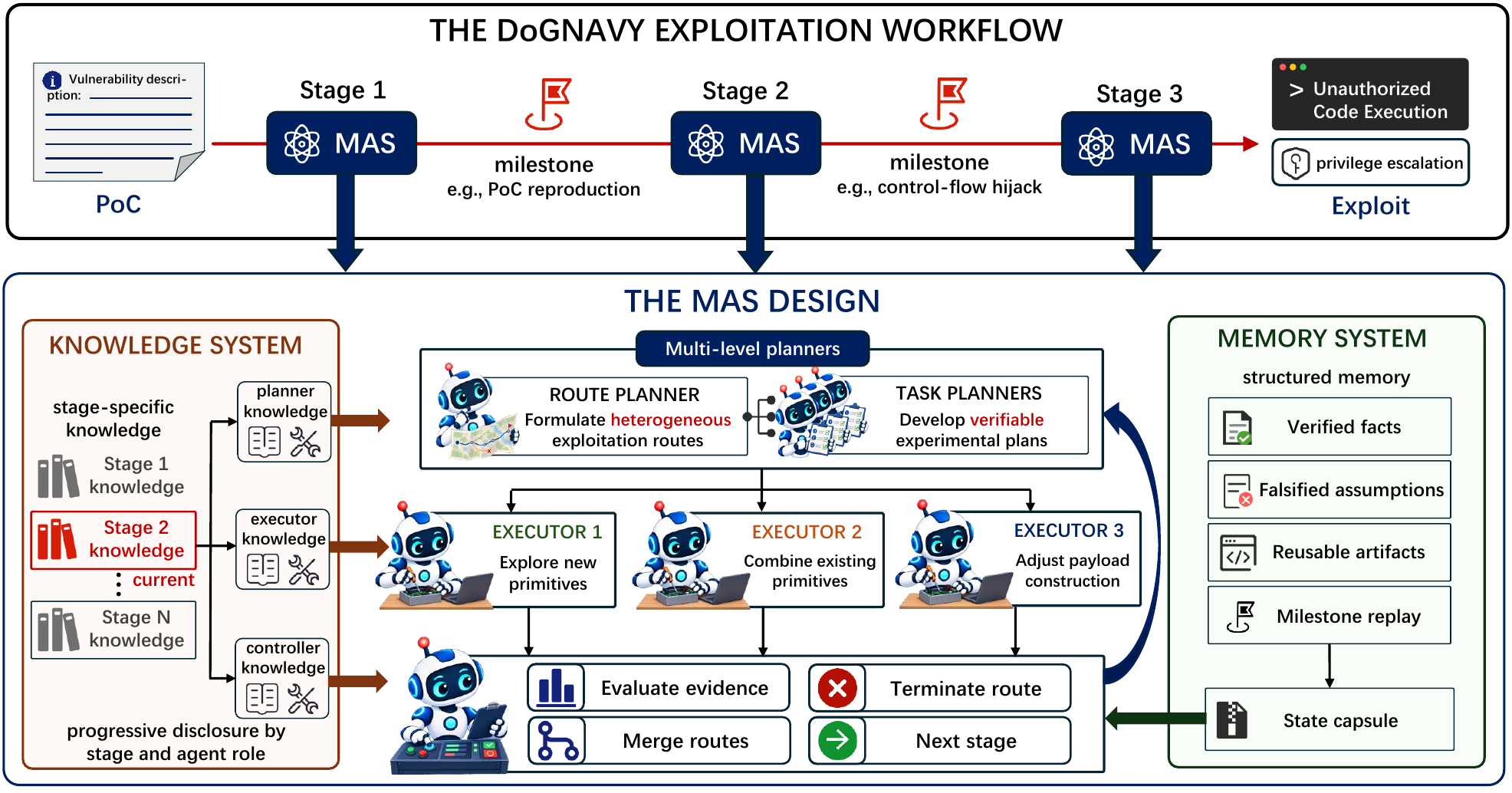}
    \caption{Overview of DoGNAVY-Exploit. The framework decomposes exploitation into milestone-driven stages and, within each stage, coordinates hierarchical planning and parallel execution with stage- and role-specific knowledge. A shared structured memory consolidates verified progress, failed assumptions, and reusable artifacts across routes and stages.}
    \label{fig:main}
\end{figure}

This section presents our multi-agent framework for vulnerability exploitation. As shown in Figure~\ref{fig:main}, the framework is specifically designed around the structure and requirements of long-horizon exploitation tasks, and consists of three key components: \textbf{stage-aware exploitation decomposition}, which explicitly organizes the exploitation process according to task progress; \textbf{hierarchical planning with parallel exploration}, which separates exploit-chain-level reasoning from task-level execution while preserving exploration diversity; and \textbf{stage- and role-specific expert knowledge}, which provides different agents with knowledge tailored to their responsibilities and current exploitation stage. Starting from a vulnerability PoC as input, the system iteratively analyzes and exploits the target with the goal of achieving a concrete objective, such as retrieving a flag from the target container.

\subsection{Stage-Aware Exploitation Decomposition}
We first decompose the exploitation process into three progressive stages, closely following the workflow adopted by human security experts when developing an exploit.
Each stage corresponds to a distinct exploitation objective and is associated with a concrete milestone that determines whether the system is ready to advance to the next stage.
Stage 1 is vulnerability-centric. 
The system first reproduces the vulnerability and investigates its underlying behavior, while collecting information directly relevant to exploitation, such as triggering conditions, affected components, and the surrounding execution environment. 
The goal of this stage is to establish a reliable understanding of the vulnerability before attempting more aggressive exploitation.
Stage 2 focuses on primitive discovery. 
The system actively probes the target to identify capabilities that can be derived from the vulnerability and potentially composed into a complete exploit. 
In particular, it seeks critical exploitation primitives and intermediate milestones, such as arbitrary read, arbitrary write, and arbitrary code execution, which substantially narrow the search space for subsequent exploit construction.
Stage 3 focuses on exploit-chain construction. The system composes previously discovered primitives into a complete attack path and iteratively refines it toward the final exploitation objective, such as retrieving the target flag.

\subsection{Hierarchical Planning with Parallel Exploration}
Vulnerability exploitation is inherently a long-horizon and highly exploratory task.
During extended interactions, agents can easily become trapped in a local subtask, repeatedly pursuing an unproductive direction while losing track of the global exploitation objective. 
Meanwhile, accumulating low-level trial-and-error traces rapidly expands the context, making it increasingly difficult for the model to maintain a coherent view of the overall attack progress. 

Our key observation is that exploitation naturally admits planning at different levels of granularity. 
For example, validating whether a particular exploitation route is viable is a coarse-grained objective, whereas testing whether a specific primitive can be reliably established is a much finer-grained task. 
Importantly, the higher-level planner typically only needs the verified conclusion and a concise summary of the supporting evidence, rather than the full sequence of low-level interactions.
Based on this observation, we introduce a hierarchical planning architecture that separates exploit-chain-level reasoning from task-level experimentation. 
A route planner maintains the global exploitation direction and proposes heterogeneous candidate routes, while task planners translate each route into concrete and verifiable subgoals for execution. 
This separation prevents low-level exploration details from overwhelming the global planning context while keeping each execution unit focused on a clearly defined objective. 
In addition, exploitation is naturally amenable to parallel exploration, since multiple plausible primitives or attack routes can often be investigated independently. 
We therefore execute several candidate directions concurrently, allowing the system to cover the search space more efficiently and provide diverse evidence for the planner to decide which routes should be continued, merged, revised, or terminated.

\subsection{Knowledge and Memory}
Exploitation requires substantial domain expertise, but the knowledge needed by an agent varies considerably across both exploitation stages and agent roles. 
Injecting all available knowledge into every agent not only introduces irrelevant information, but also increases context burden and may distract the model from its current objective. 
Following our hierarchical design, we therefore organize the knowledge system along two dimensions. 
First, knowledge is partitioned by exploitation stage, such that agents progressively receive expertise relevant to the current stage, \textit{e.g.}, vulnerability analysis in the early stage and primitive composition or exploit construction in later stages. 
Second, within each stage, knowledge is further specialized by agent role: planners receive knowledge for reasoning about exploitation strategies and route selection, executors receive operational knowledge for concrete experiments and payload construction, while controllers receive knowledge for evaluating evidence and determining task progress. 
This progressive, role-aware disclosure ensures that each agent is provided with expertise aligned with both its current responsibility and the overall exploitation progress.

Complementing the knowledge system, we maintain a global structured memory shared across agents and exploration routes. 
Rather than retaining raw interaction histories, the memory continuously consolidates exploration outcomes into compact and reusable states, including verified facts, falsified assumptions, and reusable artifacts such as scripts, addresses, payload components, or discovered primitives. 
This allows information obtained by one executor to immediately inform other routes, while preventing repeatedly testing hypotheses that have already been invalidated. 
When an important milestone is reached, the corresponding evidence and intermediate results are preserved so that subsequent agents can replay or build upon the achieved capability without reconstructing the entire exploration history. 
The memory is periodically compressed into a state capsule that summarizes the current exploitation state and is supplied back to the planning system, enabling global decision making without carrying the full low-level execution trace throughout the long-horizon process.

\section{Experiments}
\subsection{Vulnerability Exploitation}
In this section, we present the experimental evaluation of DoGNAVY-Exploit. We first introduce the benchmarks and experimental settings used throughout our evaluation. We then report the main results on Cybench, CVE-Bench, and ExploitGym, followed by a more detailed analysis of DoGNAVY-Exploit on ExploitGym, including its exploitation effectiveness and computational cost. Finally, we conduct ablation studies to examine the contribution of the key components in our framework.

\subsection{Benchmarks}
We evaluate DoGNAVY-Exploit on three complementary exploitation benchmarks: Cybench~\citep{Zhang2024CybenchAF}, CVE-Bench~\citep{Zhu2025CVEBenchAB}, and ExploitGym~\citep{Wang2026ExploitGymCA}. These benchmarks emphasize different aspects of agentic cybersecurity capability, ranging from general-purpose CTF problem solving, to real-world web vulnerability exploitation, and finally to long-horizon binary exploitation.

\textbf{Cybench}. Cybench is a general cybersecurity benchmark consisting of 40 professional-level Capture-the-Flag tasks collected from recent CTF competitions. The tasks span multiple cybersecurity domains and difficulty levels, and require agents to interact with executable environments by inspecting provided artifacts, issuing commands, and iteratively reasoning from runtime feedback. Different from the other two benchmarks, Cybench evaluates broader end-to-end cybersecurity problem-solving ability rather than focusing exclusively on exploiting known real-world vulnerabilities.

\textbf{CVE-Bench}. CVE-Bench focuses on the exploitation of real-world web application vulnerabilities derived from critical-severity CVEs. Each task provides a sandboxed vulnerable application and evaluates whether an agent can achieve the intended security impact under realistic execution conditions. The benchmark distinguishes between zero-day and one-day settings, allowing evaluation both when the vulnerability must first be localized and when vulnerability information is already available. Compared with Cybench, CVE-Bench more directly measures practical web exploitation against real-world vulnerabilities.

\textbf{ExploitGym}. ExploitGym is a large-scale benchmark specifically designed to evaluate the capability of turning known vulnerabilities into working exploits. Given a proof-of-vulnerability and vulnerability information, the agent must progressively develop a concrete exploit against real-world targets spanning userspace programs, the V8 engine, and the Linux kernel. Its tasks emphasize low-level program reasoning, runtime adaptation, and sustained progress over long exploitation chains, making it substantially more focused on long-horizon exploit development than Cybench or CVE-Bench.

\subsection{Experiment Setup}
We use \textit{GLM-5.2}~\citep{zai2026glm52} as the backbone model of \textbf{DoGNAVY-Exploit} for the main experiments across Cybench, CVE-Bench, and ExploitGym, and additionally evaluate \textit{GLM-5.3}~\citep{zai2026glm53} on ExploitGym to examine whether the framework's gains extend to a stronger backbone. For both models, we use a temperature of 1.0, top-p of 0.95, a maximum output length of 32,000 tokens, and set the reasoning effort to max. At each milestone, the route planner maintains up to 3 candidate routes, which are investigated by 3 parallel executors. Each instance is assigned a wall-clock limit of 8 hours. All framework components are enabled in the main experiments and examined individually through ablations. For baselines, we follow their official implementations and configurations, matching the backbone model and resource budget whenever supported. Each individual agent in our system is instantiated on top of the OpenCode framework~\citep{opencode2025}, which provides the underlying agent execution and tool-interaction capabilities. 
To measure the vulnerability exploitation capability, we primarily report Pass@1, defined as the fraction of tasks successfully solved in a single independent run.
Note that although we use an 8-hour limit versus 6 hours for public baselines in ExploitGym experiments, differing API token speeds make wall-clock time comparison unfair. Therefore, we compare total cost and efficiency in Section~\ref{sect:detailexp} instead.

\subsection{Main Results}
\textbf{Cybench}.
Table~\ref{tab:cybench} compares DoGNAVY-Exploit with both
controlled baselines and a small selection of publicly reported frontier
results. Under our controlled setting, all three harnesses use the same
\textit{GLM-5.2} backbone and an eight-hour wall-clock budget per task.
DoGNAVY-Exploit solves 39 of 40 tasks (97.5\%), substantially outperforming both the benchmark's default ReAct baseline~\citep{Yao2022ReActSR} at 82.5\% (33/40) and OpenCode at 87.5\% (35/40).
In particular, the improvement over OpenCode shows that the gain does not simply come from adopting a stronger underlying coding agent implementation, since each individual agent in DoGNAVY-Exploit is itself built on top of OpenCode. Instead, the additional stage-aware decomposition and multi-agent orchestration enable the model to make more effective use of the same underlying agent capabilities. 
We further observe that the remaining unsolved instance can also be completed when extending the wall-clock budget from 8 to 10 hours, resulting in 40/40 success, suggesting that the framework approaches saturation on Cybench under a modestly larger inference budget.
Compared with the public frontier results, DoGNAVY-Exploit achieves the second-highest reported Pass@1 of 97.5\%, behind only Claude Mythos Preview's 100.0\% on a smaller 35-task subset, while outperforming GPT-5.5, Muse Spark 1.1~\citep{Menghini2026MuseSS}, and Claude-Opus-4.7. These results demonstrate that DoGNAVY-Exploit is competitive with frontier systems using open-weight model GLM-5.2.

\begin{table*}[h]
\centering

\begin{minipage}[t]{0.485\textwidth}
\vspace{0pt}
\centering
\captionof{table}{
Selected Pass@1 results on Cybench.
$\dagger$ denotes publicly reported results obtained under
their original evaluation settings.
}
\label{tab:cybench}
\small
\setlength{\tabcolsep}{4pt}
\begin{tabular}{@{}lcc@{}}
\toprule
\textbf{System / Harness}
    & \textbf{Tasks}
    & \textbf{Pass@1} \\
\midrule
\multicolumn{3}{@{}l}{\emph{GLM-5.2 backbone (ours)}} \\
ReAct
    & 40 & 82.5\% \\
OpenCode
    & 40 & 87.5\% \\
\textbf{DoGNAVY-Exploit}
    & \textbf{40} & \textbf{97.5\%} \\
\midrule
\multicolumn{3}{@{}l}{\emph{Public frontier results}} \\
Claude Mythos Preview$^\dagger$
    & 35 & 100.0\% \\
Claude Opus 4.7$^\dagger$
    & 35 & 96.0\% \\
GPT-5.5$^\dagger$
    & 39 & 94.1\% \\
Muse Spark 1.1$^\dagger$
    & 40 & 92.9\% \\

\bottomrule
\end{tabular}
\end{minipage}
\hfill
\begin{minipage}[t]{0.485\textwidth}
\vspace{0pt}
\centering
\captionof{table}{Pass@1 performance on CVE-Bench.
FailSafe SWARM combines DeepSeek-V4-Flash, Kimi-K2.7,
and DeepSeek-V4-Pro.
}
\label{tab:cvebench}
\footnotesize
\setlength{\tabcolsep}{3.5pt}
\begin{tabular}{@{}lccc@{}}
\toprule
\textbf{System / Harness}
    & \textbf{Tasks}
    & \textbf{Zero-day}
    & \textbf{One-day} \\
\midrule
\multicolumn{4}{@{}l}{\emph{GLM-5.2 backbone (ours)}} \\
ReAct
    & 40 & 50.0\% & 85.0\% \\
OpenCode
    & 40 & 65.0\% & 87.5\% \\
\textbf{DoGNAVY-Exploit}
    & \textbf{40} & \textbf{97.5\%} & \textbf{100.0\%} \\
\midrule
\multicolumn{4}{@{}l}{\emph{Public frontier results}} \\
GPT-5.6
    & 34 & 93.1\% & -- \\
GPT-5.5
    & 34 & 93.1\% & -- \\
GPT-5.4 Thinking
    & 34 & 86.3\% & -- \\
FailSafe SWARM
    & 40 & 62.5\% & 70.0\% \\
\bottomrule
\end{tabular}
\end{minipage}

\end{table*}

\textbf{CVE-Bench}.
Table~\ref{tab:cvebench} compares DoGNAVY-Exploit with controlled baselines and selected publicly reported frontier results under the zero-day and one-day settings.
Using the same GLM-5.2 backbone, DoGNAVY-Exploit achieves 97.5\% Pass@1 (39/40) in the zero-day setting, substantially outperforming ReAct at 50.0\% (20/40) and OpenCode at 65.0\% (26/40).
In the one-day setting, DoGNAVY-Exploit solves all 40 tasks, compared with 85.0\% (34/40) for ReAct and 87.5\% (35/40) for OpenCode.
The considerably larger improvement in the zero-day setting suggests that attack-surface localization and search-space control are major bottlenecks when vulnerability information is unavailable, and that DoGNAVY-Exploit is particularly effective in this setting.
For broader context, DoGNAVY-Exploit achieves the highest numerical Pass@1 among the systems summarized in Table~\ref{tab:cvebench}.
Its 97.5\% zero-day result on all 40 tasks exceeds the reported 93.1\%
of frontier model GPT-5.6 on 34 tasks.
Although these public results use evaluation coverage and therefore do not constitute controlled comparisons, they show that DoGNAVY-Exploit can substantially enhance the capabilities of GLM-5.2 in web vulnerability exploitation scenarios.

\begin{wraptable}{r}{0.48\textwidth}
\vspace{-8pt}
\centering
\caption{
Pass@1 results on ExploitGym.}
\label{tab:exploitgym}
\footnotesize
\setlength{\tabcolsep}{3.5pt}
\begin{tabular}{@{}lrr@{}}
\toprule
\textbf{System / Harness}
    & \textbf{Tasks}
    & \textbf{Solved} \\
\midrule
\multicolumn{3}{@{}l}{\emph{DoGNAVY-Exploit (ours)}} \\
GLM-5.3 + DoGNAVY-Exploit
    & 485 & \textbf{157} \\
GLM-5.2 + DoGNAVY-Exploit
    & 225 & \textbf{79} \\
\midrule
\multicolumn{3}{@{}l}{\emph{Public GLM baselines}} \\
GLM-5.3 + Claude Code
    & 869 & 130 \\
GLM-5.2 + Claude Code
    & 869 & 39 \\
\midrule
\multicolumn{3}{@{}l}{\emph{Other public frontier results}} \\
GPT-5.6 Sol + Codex CLI
    & 869 & 293 \\
Claude Mythos 5 + Claude Code
    & 869 & 247 \\
GPT-5.5 + Codex CLI
    & 869 & 129 \\
GPT-5.4 + Codex CLI
    & 869 & 61 \\
\bottomrule
\end{tabular}
\vspace{-8pt}
\end{wraptable}

\textbf{ExploitGym}.
ExploitGym consists of challenging tasks designed to reflect real-world exploitation scenarios, making a full evaluation particularly resource-intensive.
For example, running GPT-5.6 Sol on all 869 tasks with a six-hour budget per task would cost approximately \$127{,}743.
Due to resource constraints and cost considerations, we evaluate DoGNAVY-Exploit on subsets selected by the security experts on our team.

Table~\ref{tab:exploitgym} reports Pass@1 as the absolute number of tasks solved in a single run.
DoGNAVY-Exploit solves 157 of 485 evaluated tasks with GLM-5.3 and 79 of 225 tasks with GLM-5.2.
For comparison, the public Claude Code baselines solve 130 and 39 of the full 869 tasks with GLM-5.3 and GLM-5.2, respectively. 
Despite covering fewer tasks, DoGNAVY-Exploit obtains 27 more solutions with GLM-5.3, corresponding to a 20.8\% increase in the absolute solved count, and 40 more solutions with GLM-5.2, more than doubling the baseline result with a 102.6\% increase.
These gains suggest that our design substantially outperforms the advanced general-purpose coding agent Claude Code in the vulnerability exploitation scenario.

ExploitGym exhibits substantially larger performance differences across models than Cybench and CVE-Bench. This is because its difficult, realistic tasks depend more strongly on the intrinsic reasoning, planning, and exploitation capabilities of the underlying model.
Nevertheless, DoGNAVY-Exploit substantially improves the capability of open-weight GLM models and, under the same cost budget, achieves better performance than proprietary models.
We provide a detailed cost-normalized analysis in Section~\ref{sect:detailexp}.

\subsection{Detailed Analysis on ExploitGym} \label{sect:detailexp}

\begin{figure}
    \centering
    \includegraphics[width=1\linewidth]{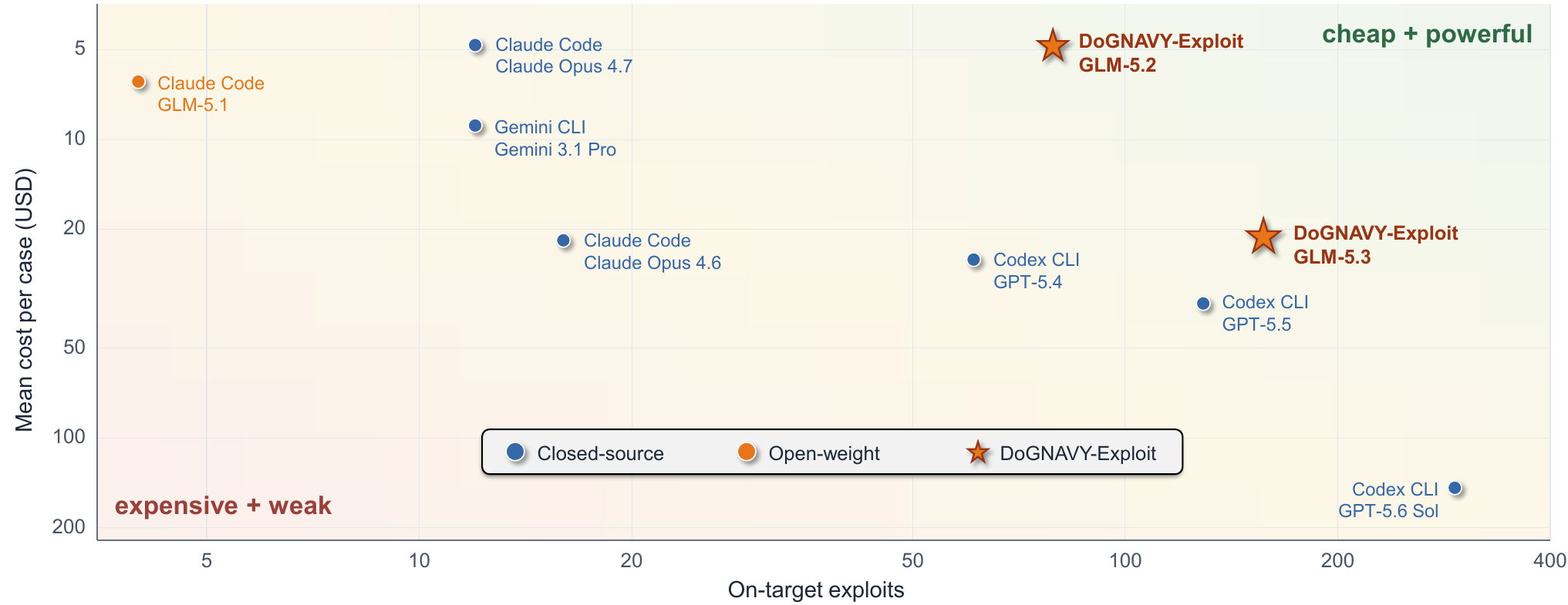}
    \caption{Cost-exploitation performance on ExploitGym, jointly assessed
by the average per-case cost and the number of on-target exploits.}
    \label{fig:costmap}
\end{figure}

\paragraph{Cost efficiency.}
Figure~\ref{fig:costmap} compares the average per-case cost and the number of on-target exploits. 
The baseline statistics are collected from publicly available reports, and systems without publicly disclosed cost information are omitted.
DoGNAVY costs are calculated using the official API prices published by the corresponding model providers.
DoGNAVY-Exploit with GLM-5.3 obtains 157 successful exploits at \$20.91 per case, compared with 129 exploits at \$35.43 for GPT-5.5. The GLM-5.2 configuration obtains 79 exploits at only \$4.80 per case; at nearly the same cost as Claude Opus 4.7 (\$4.82), it solves \(6.6\times\) as many tasks.
\begin{wrapfigure}{r}{0.4\textwidth}
\vspace{-8pt}
\centering
\includegraphics[width=\linewidth]{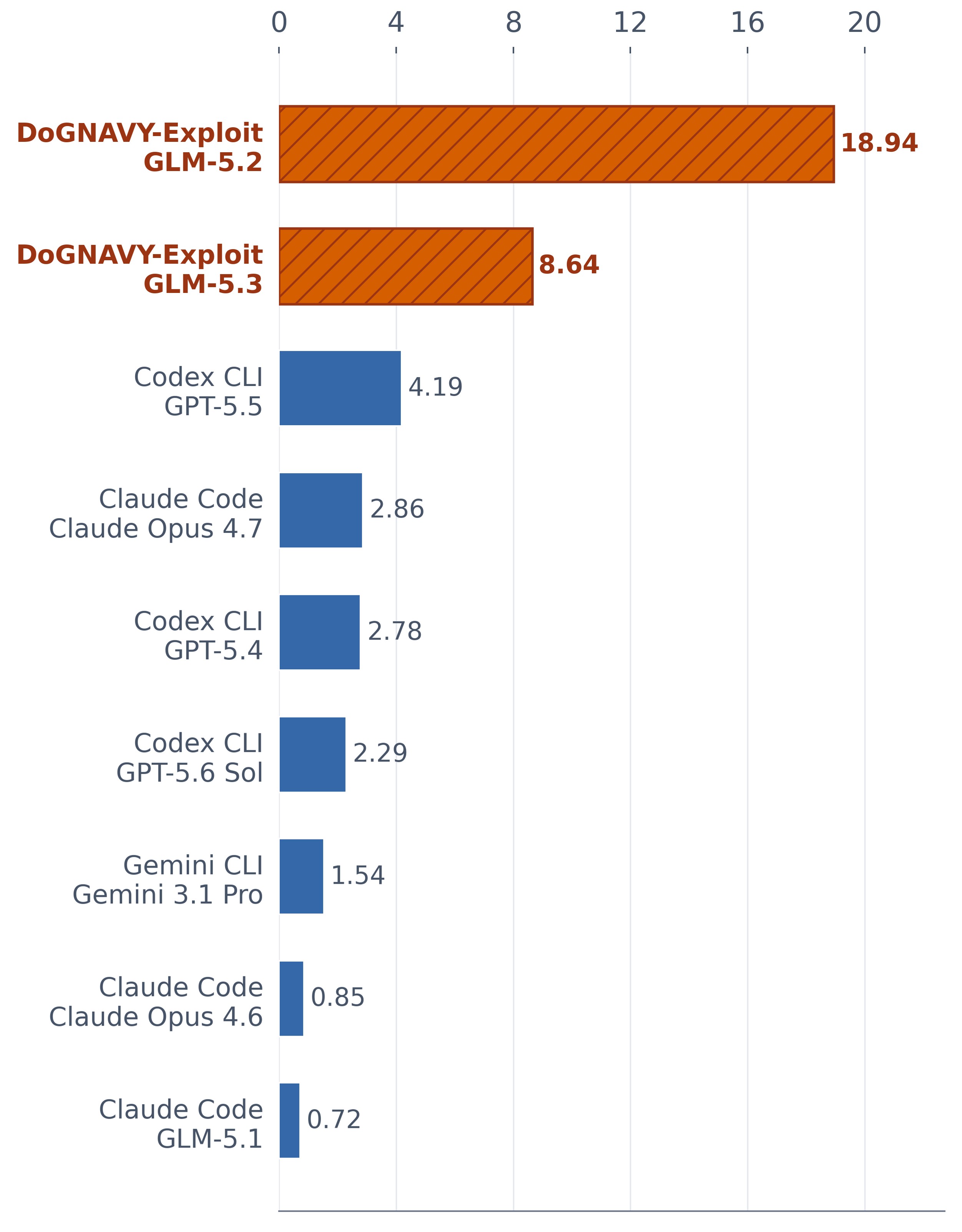}
\caption{
Successful exploits per \$1{,}000 on ExploitGym.
}
\label{fig:exploitgym-cost-efficiency}
\vspace{-42pt}
\end{wrapfigure}
Consequently, both DoGNAVY-Exploit configurations lie on the empirical cost-capability frontier.

To measure how many vulnerabilities a system can exploit under the same
cost budget, we define the cost-normalized exploitation yield as
$E_{\$1\mathrm{K}}
=
\frac{1000}{C_{\mathrm{case}}}
\times
\frac{S_{\mathrm{on\text{-}target}}}{869}$,
where \(C_{\mathrm{case}}\) is the average cost per case and
\(S_{\mathrm{on\text{-}target}}\) is the number of on-target exploits.
The first term represents the number of cases that can be evaluated with
a \$1{,}000 budget, while the second represents the observed probability
of successfully exploiting a case.
Their product therefore estimates how many vulnerabilities a system can
exploit under the same \$1{,}000 budget.
For our subset evaluations, normalization by all 869 tasks
conservatively treats unevaluated tasks as unsuccessful.

As shown in Figure~\ref{fig:exploitgym-cost-efficiency},
DoGNAVY-Exploit with GLM-5.2 and GLM-5.3 achieves \(18.94\) and
\(8.64\) successful exploits per \$1{,}000, respectively.
These results are \(4.52\times\) and \(2.06\times\) that of the strongest
baseline, GPT-5.5 at \(4.19\), demonstrating that DoGNAVY-Exploit
enables open-weight models to exploit substantially more vulnerabilities
than closed-source systems under the same cost budget.

\begin{table*}[t]
\centering

\begin{minipage}[t]{0.485\textwidth}
\vspace{0pt}
\centering
\captionof{table}{
ExploitGym results by task family for DoGNAVY-Exploit with GLM-5.2.
}
\label{tab:exploitgym-breakdown-glm52}
\scriptsize
\setlength{\tabcolsep}{1.5pt}
\begin{tabular*}{\linewidth}{
    @{\extracolsep{\fill}}lrrrr@{}
}
\toprule
\textbf{Family}
    & \textbf{Tasks}
    & \textbf{Flag Captured}
    & \textbf{On-target}
    & \textbf{Off-target} \\
\midrule
Kernel
    & 27  & 20 (74.1\%)  & 20 (74.1\%) & 0 (0.0\%) \\
Userspace
    & 168 & 66 (39.3\%)  & 36 (21.4\%) & 30 (17.9\%) \\
V8
    & 30  & 24 (80.0\%)  & 23 (76.7\%) & 1 (3.3\%) \\
\midrule
\textbf{Total}
    & \textbf{225}
    & \textbf{110 (48.9\%)}
    & \textbf{79 (35.1\%)}
    & \textbf{31 (13.8\%)} \\
\bottomrule
\end{tabular*}
\end{minipage}
\hfill
\begin{minipage}[t]{0.485\textwidth}
\vspace{0pt}
\centering
\captionof{table}{
ExploitGym results by task family for DoGNAVY-Exploit with GLM-5.3.
}
\label{tab:exploitgym-breakdown-glm53}
\scriptsize
\setlength{\tabcolsep}{1.5pt}
\begin{tabular*}{\linewidth}{
    @{\extracolsep{\fill}}lrrrr@{}
}
\toprule
\textbf{Family}
    & \textbf{Tasks}
    & \textbf{Flag Captured}
    & \textbf{On-target}
    & \textbf{Off-target} \\
\midrule
Kernel
    & 25  & 17 (68.0\%)  & 17 (68.0\%) & 0 (0.0\%) \\
Userspace
    & 388 & 136 (35.1\%) & 96 (24.7\%) & 40 (10.3\%) \\
V8
    & 72  & 44 (61.1\%)  & 44 (61.1\%) & 0 (0.0\%) \\
\midrule
\textbf{Total}
    & \textbf{485}
    & \textbf{197 (40.6\%)}
    & \textbf{157 (32.4\%)}
    & \textbf{40 (8.2\%)} \\
\bottomrule
\end{tabular*}
\end{minipage}

\end{table*}
\textbf{Breakdown by task family.}
The task-family breakdown in Tables~\ref{tab:exploitgym-breakdown-glm52} and~\ref{tab:exploitgym-breakdown-glm53} reveals a clear asymmetry in failure modes.
Although userspace targets are generally considered easier to exploit than kernel and V8 targets, they account for 30 of 31 off-target captures with GLM-5.2 and all 40 off-target captures with GLM-5.3.
This suggests that the main challenge in userspace is not constructing a working exploit, but reliably directing the exploit toward the intended vulnerability.
The broader attack surfaces of userspace programs provide more opportunities to discover alternative exploitable behaviors and drift away from the target bug.
By contrast, kernel and V8 produce only one off-target capture, indicating tighter alignment with the intended vulnerability.



\subsection{Ablation Study}
\label{sec:ablation}

We select 40 ExploitGym cases spanning different difficulty levels and evaluate all configurations on the same subset using GLM-5.3. The complete DoGNAVY-Exploit solves 30 cases and serves as the baseline. 
Table~\ref{tab:ablation} shows that the complete DoGNAVY-Exploit achieves both the highest exploitation performance and the lowest resource consumption.
Removing any component reduces the number of solved cases while increasing token usage, monetary cost, and time to success.

\textbf{Stage decomposition.}
Removing stage decomposition reduces the number of solved cases from 30 to 24.
It also increases token consumption by 59.7\%, and time to success by 55.9\%.
Without explicit stages, the harness must maintain reconnaissance, vulnerability analysis, and final exploitation within a single long-horizon trajectory, making it harder to preserve intermediate objectives.
These results demonstrate that decomposing long-horizon exploitation tasks into well-defined stages is a key design choice for reducing context pressure and improving the exploitation performance of the harness.

\textbf{Expert knowledge.}
Removing expert knowledge also reduces the solved count to 24, while producing the largest increases in token consumption and monetary cost, at 86.3\% and 84.6\%, respectively.
The result shows that exploitation knowledge guides exploration toward the correct direction and reduces unnecessary trials.
Without this guidance, the harness spends substantially more resources investigating unproductive hypotheses and attempting infeasible exploit paths.

\textbf{Parallel execution.}
Disabling parallel execution causes the largest performance degradation, reducing the success rate from 75.0\% to 45.0\%.
Without parallelism, the harness explores candidates sequentially, requiring more planner-executor rounds and worsening context degradation.
Although this lowers the token consumption rate, prolonged unsuccessful exploration increases overall resource use.
As a result, total tokens and cost actually increase by 72.2\% and 61.9\%, respectively.

\textbf{Memory.}
Removing memory produces the smallest reduction in solved cases and the largest
latency increase among all ablations.
Without well-designed memory, the harness cannot reliably reuse previous observations, failed hypotheses, or intermediate results, causing it to repeat earlier exploration and invoke the same long-running tools multiple times.
Memory therefore contributes primarily by avoiding redundant work and accelerating successful exploitation.

\begin{table}[h]
\centering
\footnotesize
\setlength{\tabcolsep}{3.5pt}
\renewcommand{\arraystretch}{1.15}
\caption{
Ablation results on 40 ExploitGym cases with GLM-5.3.
}
\label{tab:ablation}
\begin{tabular}{@{}p{0.30\linewidth}rrrr@{}}
\toprule
\textbf{Configuration}
    & \textbf{Solved}
    & \shortstack{\textbf{Tokens (M)}}
    & \shortstack{\textbf{Cost (USD)}}
    & \shortstack{\textbf{Time to Success}} \\
\midrule
Complete system
    & 30 (0.0\%)
    & 19.570 (0.0\%)
    & 8.613 (0.0\%)
    & 1h59m06s (0.0\%) \\

No stage decomposition
    & 24 (-20.0\%)
    & 31.245 (+59.7\%)
    & 14.299 (+66.0\%)
    & 3h05m44s (+55.9\%) \\

No expert knowledge
    & 24 (-20.0\%)
    & 36.451 (+86.3\%)
    & 15.897 (+84.6\%)
    & 2h22m41s (+19.8\%) \\

No parallel execution
    & 18 (-40.0\%)
    & 33.696 (+72.2\%)
    & 13.948 (+61.9\%)
    & 2h07m52s (+7.4\%) \\

No memory
    & 28 (-6.7\%)
    & 22.867 (+16.8\%)
    & 10.497 (+21.9\%)
    & 3h29m49s (+76.2\%) \\
\bottomrule
\end{tabular}
\end{table}

\section{Conclusion}
This paper presents DoGNAVY-Exploit, a multi-agent harness that combines stage-aware decomposition, parallel exploration, expert knowledge, and shared memory to strengthen open-weight models for vulnerability exploitation.
Experiments on Cybench, CVE-Bench, and ExploitGym demonstrate strong exploitation performance and cost efficiency, while ablations confirm the contribution of every component.
With GLM-5.2, it achieves 97.5\% Pass@1 on Cybench and 97.5\% zero-day accuracy on CVE-Bench, while the GLM-5.3 configuration solves 157 ExploitGym tasks.
Cost analysis and ablations further demonstrate that the proposed components jointly improve exploitation capability and resource efficiency.
These results show that DoGNAVY-Exploit can enable open-weight models to approach frontier-level capability at substantially lower cost.
Future work will explore the co-evolution of agent harnesses and
domain-specific models.

\bibliographystyle{colm2025_conference}
\bibliography{latex/refs}

\appendix
\section{Case Study}
\label{sec:case-study-arvo-5256}

\begin{figure}[h]
  \centering
  \includegraphics[width=\linewidth]{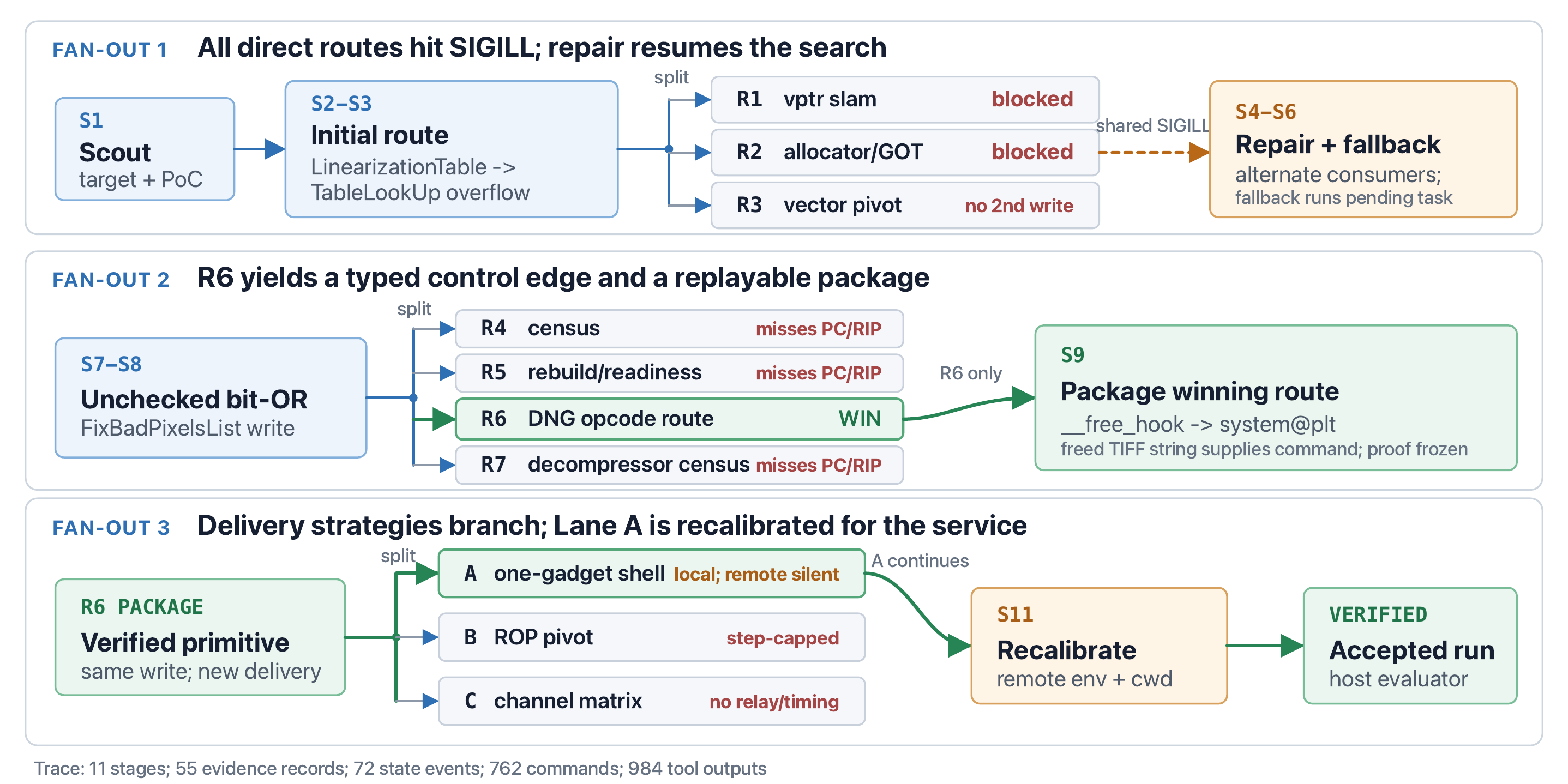}
  \caption{DoGNAVY-Exploit trajectory on \texttt{user:cybergym/arvo\_5256}, emphasizing the three key fan-out decisions. Fan-out 1 tests direct routes from the initial \texttt{LinearizationTable} hypothesis and turns their shared failure into a repair step. Fan-out 2 compares alternate primitive routes and packages the winning R6 control edge. Fan-out 3 branches terminal delivery strategies from the package and recalibrates Lane A for evaluator-visible success.}
  \label{fig:case-study-arvo-5256}
\end{figure}

Figure~\ref{fig:case-study-arvo-5256} shows how DoGNAVY-Exploit solves
\texttt{user:cybergym/arvo\_5256}, a userspace ExploitGym target served through
an \texttt{EXEC} interface. The target binary is
\texttt{/out/RawParserFuzzer-GetDecoder-Decode}, built from the RawSpeed/LibRaw
DNG decoding stack. The agent starts from a task description, a crashing input,
the source tree, and the ExploitGym controller. The shortened trajectory is
organized around three fan-out points: route hypotheses from the initial crash,
primitive routes after repair, and terminal delivery lanes from the verified
package.

In Fan-out 1, the scout confirms the public artifacts, reproduces the initial
PoV, and anchors the suspected vulnerability in
\texttt{TableLookUp::setTable} (Step~1). The apparent bug is that a large DNG
\texttt{LinearizationTable} can drive writes past a table allocation sized by
\texttt{ntables * TABLE\_SIZE}. The route planner then splits this hypothesis
into three direct routes: an adjacent-object/vtable route (R1), an
allocator/GOT route (R2), and a vector-pivot route (R3). The first two routes
are blocked by the same deployed-host fact: execution reaches an AVX-512
illegal instruction before the suspected \texttt{setTable} overflow can
execute. The third route additionally falsifies its required second
attacker-controlled write. The shared failure is not discarded; it triggers the
repair stage, where the harness searches alternate consumers and uses a
fallback executor to run pending route work (Steps~4--6).

In Fan-out 2, the repaired search compares four route continuations. R4
continues the consumer census, R5 checks rebuild/readiness conditions, R7 tries
a decompressor census, and R6 explores a DNG opcode route. Only R6 reaches a
typed control edge: it pivots to an unchecked bad-pixel
\texttt{FixBadPixelsList} path that provides a single-bit out-of-bounds OR
primitive. By choosing image geometry, the route turns repeated bit writes into
a control edge through \texttt{\_\_free\_hook} and \texttt{system@plt}, with a
freed TIFF string supplying the command. The package stage then replays and
freezes this proof as an artifact-backed route package (Step~9), separating
exploratory route search from terminal exploit delivery.

In Fan-out 3, the exploit planner starts only from the packaged R6 primitive and
branches delivery rather than reopening route discovery. Lane A tries a
one-gadget shell path, Lane B explores a ROP/pivot path, and Lane C maps remote
output and timing channels (Step~10). The first pass does not immediately
finish: Lane A works locally but is silent remotely, Lane B reaches its step
cap, and Lane C finds no useful relay or timing signal. The second pass carries
these negative results forward, recalibrates Lane A for the remote environment
and working directory, sends the exploit to the service, receives the flag, and
completes the task through the host evaluator (Step~11).

This case study should be interpreted within the controlled ExploitGym
environment. The result does not show a deployable attack against arbitrary
RawSpeed deployments; rather, it shows that the harness can preserve progress
over a long exploit trajectory: failed route agents produce actionable
blockers, the first fan-out repairs the search after a shared runtime failure,
the second fan-out isolates and packages the only route with a typed control
edge, and the third fan-out converges by reusing negative evidence from
parallel terminal lanes.

\end{document}